\documentclass[runningheads]{llncs}

\usepackage{eccv}

\usepackage{eccvabbrv}

\usepackage{graphicx}
\usepackage{booktabs}

\usepackage[accsupp]{axessibility}  

\usepackage{hyperref}

\usepackage{orcidlink}

\usepackage{multirow}
\usepackage{multicol}
\usepackage{caption}
\usepackage{color, colortbl}
\definecolor{Gray}{gray}{0.9}
\definecolor{green}{RGB}{0, 133, 21}
\definecolor{red}{rgb}{0.8,0,0}  
\definecolor{backcolor}{RGB}{232, 242, 255}
\makeatletter
\def\blfootnote{\xdef\@thefnmark{}\@footnotetext}

\begin{document}

\title{Implicit Concept Removal of Diffusion Models} 


\author{Zhili Liu\inst{1,2}\thanks{Equal contribution. Contact:
\texttt{zhili.liu@connect.ust.hk}} \and
Kai Chen\inst{1\star} \and
Yifan Zhang\inst{3} \and
Jianhua Han\inst{2} \and
Lanqing Hong\inst{2} \and \\
Hang Xu\inst{2} \and
Zhenguo Li\inst{2} \and
Dit-Yan Yeung\inst{1} \and
James T. Kwok\inst{1}
}

\authorrunning{Z.~Liu et al.}

\institute{
Hong Kong University of Science and Technology \and
Huawei Noah’s Ark Lab \quad $^{3}$\ National University of Singapore \\
Project Page: \url{https://kaichen1998.github.io/projects/geom-erasing/}
}

\maketitle

\newcommand{\methodname}{\textsc{Geom-Erasing}\xspace}

\vspace{-5mm}
\begin{abstract}
Text-to-image (T2I) diffusion models often inadvertently generate unwanted concepts such as watermarks and unsafe images. 
These concepts, termed as the ``implicit concepts'', could be unintentionally learned during training and then be generated uncontrollably during inference.
Existing removal methods still struggle to eliminate implicit concepts primarily due to their dependency on the model's ability to recognize concepts it actually can not discern.
To address this, we utilize the intrinsic geometric characteristics of implicit concepts and present the \methodname, a novel concept removal method based on the geometric-driven control.
Specifically, once an unwanted implicit concept is identified, we integrate the existence and geometric information of the concept into the text prompts with the help of an accessible classifier or detector model.
Subsequently, the model is optimized to identify and disentangle this information, which is then adopted as negative prompts during generation.
Moreover, we introduce the \textit{Implicit Concept Dataset} (\textbf{ICD}), a novel image-text dataset imbued with three typical implicit concepts (\ie, QR codes, watermarks, and text), reflecting real-life situations where implicit concepts are easily injected.
\methodname effectively mitigates the generation of implicit concepts, achieving the state-of-the-art results on the Inappropriate Image Prompts (I2P) and our challenging Implicit Concept Dataset (ICD) benchmarks.
  \keywords{Concept Erasure \and Implicit Concept \and Geometric Control}
\end{abstract}

\begin{figure}[t]
\centering
\includegraphics[width=\linewidth]{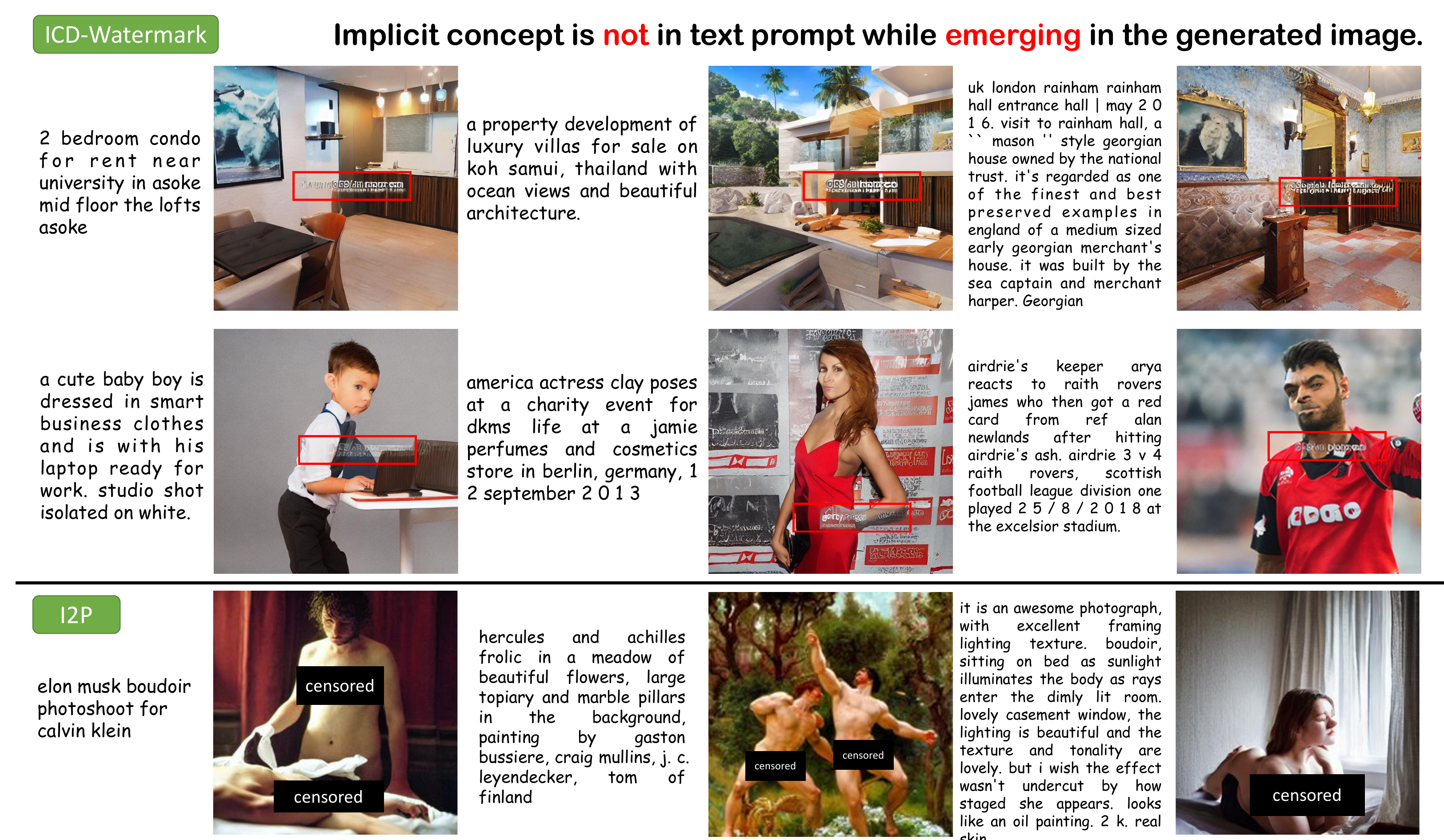}
  \vspace{-3mm}
  \caption{\textbf{Generation of implicit concepts.} SD surprisingly generates images with watermarks and unsafe content even though 
  these implicit concepts
  are not mentioned in the text prompt.
  }
  \vspace{-3mm}
  \label{fig:intro}
\end{figure}

\vspace{-6mm}
\section{Introduction}
\label{sec:intro}

Text-to-image diffusion models (DMs) have become increasingly popular due to their exceptional proficiency in generating high-quality images~\cite{ho2020denoising,zhang2023expanding,li2023trackdiffusion,gao2023magicdrive}. 
Despite the advancements, DMs sometimes inadvertently generate unwanted \textit{concepts} such as watermarks~\cite{lee2023holistic} or unsafe images~\cite{schramowski2023safe}.
Take Stable Diffusion (SD)\footnote{https://huggingface.co/runwayml/stable-diffusion-v1-5}~\cite{rombach2022high} as an example. 
When text prompts on topics related to indoor furniture or human portrait are used, 
surprisingly we  find that 
the generated images often contain watermarks 
(Fig.~\ref{fig:intro}),
even though 
no watermark-related keyword 
is mentioned 
in the text prompts.
In addition, SD is more likely to generate unsafe content when the text prompts contain art and women~\cite{schramowski2023safe}. Evaluations on
the I2P dataset~\cite{schramowski2023safe} and
our constructed ICD-Watermark dataset (specified in Sec.~\ref{sec:setup})
show that
watermarks 
and unsafe content could
show up in 11\% 
and 39\% of the generated images, respectively.

In this paper, we define \textbf{implicit concepts (IC)} as \textit{concepts that are not explicitly specified in the text prompts but are still generated by the DMs.} 
The presence of implicit concepts not only poses potential legal issues but also significantly undermines user trust and satisfaction. 
Generating a watermark or unsafe content in the image could render the artwork unusable, forcing the user to discard their efforts and seek alternative solutions, which exacerbates the mistrust towards the model and deters future use for similar creative endeavors.

Avoiding the generation of implicit concepts is difficult.
Even when the DM is 
trained on 
datasets that are supposed to be
watermark-free or not-safe-for-work (NSFW) filtered 
\cite{rombach2022high,dalle2bias},
implicit concepts might still be generated~\cite{lee2023holistic,brack2023mitigating} due to inherent limitations in detecting and filtering all problematic images from the web-crawled datasets~\cite{birhane2021multimodal}. 
Our research instead considers the alternative prevalent \textit{post-hoc} strategy that erases undesirable concepts from the pre-trained DMs.
This encompasses methods that fine-tune DMs on paired images (one containing the concepts for erasure and the other does not), to redirect the generation away from specific concepts~\cite{gandikota2023erasing, kumari2023conceptablation}, or reduce activation values in the cross-attention module~\cite{zhang2023forgetmenot}. 
Other methods carefully design the text prompts~\cite{ho2022classifier} and diffusion process~\cite{schramowski2023safe} to navigate the model negatively away from the unwanted concepts. 
While these strategies are effective for erasing explicit concepts such as semantic objects and art styles, we find that they do not perform well on erasing implicit concepts, as quantitatively shown in Table~\ref{tab:sota_setting1} and Table~\ref{tab:sota}.

Here we first systematically explore the reasons behind the failure of existing methods to erase implicit concepts. In particular, we identify a \textit{core mismatch}, in that these methods assume the concept to be erased can be controllably generated or recognized by the DM, which is not feasible for implicit concepts.
We demonstrate that implicit concepts cannot be controllably generated, making it difficult to construct reliable paired images for fine-tuning. Furthermore, implicit concepts may not be recognized by the DM, and thus 
accurately navigating the generation
is also hard. 
Refer to Sec.~\ref{sec:preli} for detailed experiments.

Next,
we study why DMs generate concepts they are not aware of.
We conjecture that it stems from the training dataset which contains \textit{images with these concepts but the text conditions do not}. 
To demonstrate this,
we construct an \textbf{Implicit Concept Dataset (ICD)}, containing three implicit concepts 
(QR codes, watermarks, and text) that commonly exist in 
real-world 
image databases. 
Training on the ICD shows a high chance of generating images containing these unwanted implicit concepts, while the resulting model is unaware of them, as detailed in Appendix~\ref{sec:app_preli}.
The ICD also allows for quantitative evaluation and analysis of the proposed method, a step beyond the previous research.

Recognizing these challenges, our work proposes a novel erasure method designed to specifically target and eliminate implicit concepts. Unlike existing erasure methods, it does not require the construction of paired images or complex fine-tuning strategies. 
Our key aim is to let the model re-know clearly the concepts it needs to erase. We observe a consistent feature where these concepts 
often exists \textbf{in certain parts} of the image. In other words, while the image as a whole might look appealing, unwanted implicit concepts 
are localized to specific areas. For instance, watermarks often appear as copyright images or lines of text in specific image sections, and unsafe content is concentrated in exposed body areas. Based on these insights, we propose \methodname, a straightforward yet effective technique aiming at removing implicit concepts in diffusion models.
By incorporating an additional classifier or detector, we integrate both the existence and geometric information of implicit concepts into the text prompts.  
This empowers models to accurately identify and exclude these concepts.
Notably, \methodname converts this information into text prompts without accessing the parmeters of the classifier or detector. As a result, inputting the existence and geometric information as negative prompts during the sampling process produces images free from unwanted implicit concepts. Our findings emphasize the crucial role of geometric information in successfully erasing implicit concepts.

For performance evaluation,
we use
two settings
mimicking real-world scenarios:
1) \textbf{Model Removal}:
The pre-trained DM \textbf{already} contains watermark and unsafe concepts, and we aim to remove these implicit concepts 
without harming the generation quality of other concepts.
2) \textbf{Data Removal}: The users may \textbf{fine-tune} the DM with datasets containing implicit concepts. 
Under both settings, we successfully reduce the chance of generating implicit concepts on ICD, and outperform previous state-of-the-art for eliminating the unsafe content on the Inappropriate Image Prompts (I2P)~\cite{schramowski2023safe} benchmark.
\\
\\
Our contributions can be summarized as:
\begin{enumerate}
\item We 
present the problem of eliminating implicit concepts (IC), and uncover a fundamental shortcoming of current erasure techniques, namely that they assume that concepts can be intentionally generated or recognized by DMs, an assumption not feasible for implicit concepts, leading to their failure in eradicating these concepts.
\item We construct the \textbf{Implicit Concept Dataset}, containing three sub-datasets embedded with varied implicit concepts, to serve as substantial resources to propel future research endeavors to resolve the problem.
\item We propose \methodname, a novel concept-removal method verified through two different settings: \textit{model removal}, and \textit{data removal}, demonstrating its robust capability to eliminate implicit concepts, proved effective in real-world applications.
\end{enumerate}

\section{Related work}
\label{sec:related_work}

\textbf{Diffusion models}~\cite{sohl2015deep, ho2020denoising}
excel in various generative tasks such as density estimation~\cite{kingma2021variational}, image synthesis~\cite{dhariwal2021diffusion}, text-to-image generation~\cite{rombach2022high, balaji2022ediffi, saharia2022photorealistic,zhang2023hipa}, and so on. It transforms a data distribution to a Gaussian distribution by incrementally injecting noise and subsequently reversing this process through denoising to reconstruct the original distribution. This study particularly concentrates on text-to-image generation using diffusion models that are pre-trained on extensive datasets~\cite{rombach2022high}. Such models, while capable of generating diverse content based on text conditions, also present notable risks such as generating harmful~\cite{schramowski2023safe}, watermarked, and content infringing on copyright~\cite{zhang2023forgetmenot}. Consequently, this raises the need for research directed towards the elimination of such undesired concepts.

\vspace{+3mm}
\noindent\textbf{Concept erasing in diffusion models.}
Current methods to erase unwanted concepts mainly depend on the model's ability to recognize those concepts. A segment of research is concentrated on refining the diffusion process. Techniques such as Negative Prompt (NP)~\cite{ho2022classifier} and Safe Latent Diffusion (SLD)~\cite{schramowski2023safe, brack2023sega} use well-designed negative prompts. They employ enhanced Classifier-free guidance for more refined control, steering diffusion models away from generating specific, undesirable concepts. These approaches depend heavily on the model’s pre-trained understanding of the concept. Another approach is to fine-tune the model to remove specific concepts. For instance, Erased Stable Diffusion (ESD)~\cite{gandikota2023erasing} generates images with an unwanted concept and then guides the model away from creating such content. Forget-Me-Not (FMN)~\cite{zhang2023forgetmenot} utilizes textual inversion to bolster the model's recognition of the existence of the specific concept, subsequently adjusting the cross-attention~\cite{vaswani2017attention} scores between undesired concept and image content, resulting in images exhibiting diminished response to undesired concepts. However, we discovered that just adding existence information to the model is not enough to remove implicit concepts completely. So, we also include geometric information to the model. This helps reduce the appearance of unwanted concepts significantly.

\section{Preliminary Study}
\label{sec:preli}
Section~\ref{sec:statement}
first introduces the definition of the implicit concept removal. 
In Section~\ref{sec:pre_pre},
we perform experiments to demonstrate that implicit concepts cannot be generated in a controlled manner and are also not identifiable by Stable Diffusion (SD). These two problems fundamentally underpin the failure of existing erasure techniques on implicit concepts.

\subsection{Problem Statement}
\label{sec:statement}

This work addresses the challenge of eliminating unwanted and unintended implicit concepts from diffusion models. In  particular, we focus on the Latent Diffusion Model~\cite{rombach2022high} (Stable Diffusion (SD) specifically).
We study two realistic settings: 
(i) \textbf{Model removal}: 
We target at the removal of implicit concepts 
already presenting in SD, without compromising the original quality of generation. Here, we begin by creating image-text data, denoted $\mathcal{D} = \left\{X, Y\right\}$.
The images are generated by the diffusion model containing implicit concepts $y_{im}$.
We aim to fine-tune a model that eliminates the generation of IC while preserving its original capabilities.
(ii) \textbf{Data removal}: We consider situations where users need to fine-tune the SD on a personalized dataset $\mathcal{D}$, and this dataset contains implicit concept $y_{im}$. 
Our objective is to fine-tune a model, to generate images that closely resemble $\mathcal{D}$ but with the implicit concept removed.

Following SD, we train diffusion models in the latent space of VQ-VAE~\cite{VQVAE}. An encoder $\mathcal{E}$ maps an image $x \in X$ to the latent code $z=\mathcal{E}(x)$, and the decoder 
$D$ then
reconstructs the image as $D(z) = x$. During fine-tuning, the diffusion model optimizes a UNet~\cite{dhariwal2021diffusion,ho2020denoising,ronneberger2015u,song2020score} to predict the unscaled noise added to the latent code of the image. The loss function is:
\begin{equation}
    \mathcal{L}_{SD} = \mathbb{E}_{z_t\sim \mathcal{E}_t(x),  y\sim Y, \epsilon \sim \mathcal{N}(0,1), t} \left[ 
    \left \| \epsilon - \epsilon_{\theta}(z_t, t, c_\theta(y)) \right \|_2^2
    \right],
\label{equ:sd_loss}
\end{equation}
where $y\sim Y$ is the input text, $t$ is the time step, $z_t$ is the noised latent code of the image, $\epsilon$ is an unscaled noise sample, and $\epsilon_\theta$ is the denoising network to be fine-tuned. 
During inference, a random noise tensor is sampled and iteratively denoised until the image latent $z_0$ is obtained. 
The image is then generated by the decoder as $x' = D(z_0)$.

\begin{figure}[t]
  \centering
  \begin{subfigure}[t]{1\textwidth}
      \centering
      \includegraphics[width=0.8\textwidth]{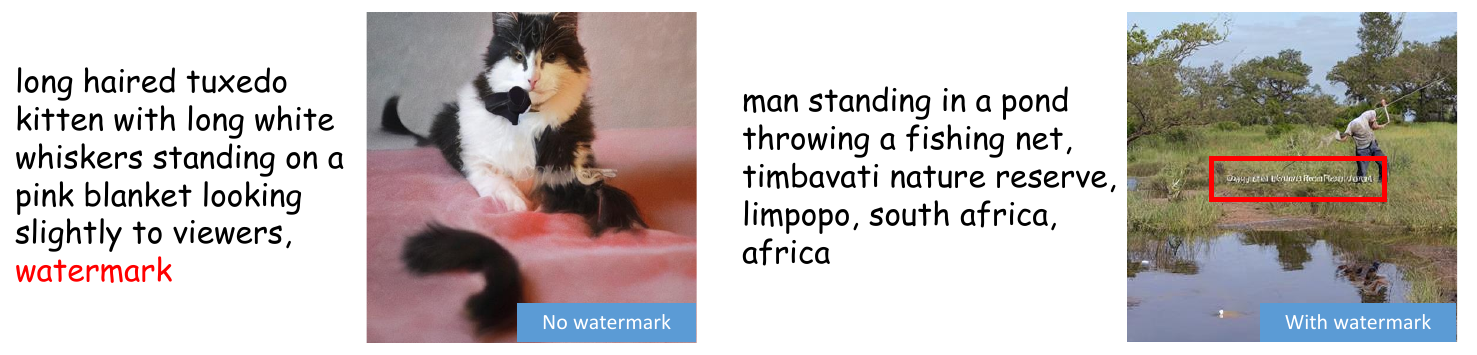}
      \caption{Including "watermark" in text prompt does not ensure its appearance in image, and vice versa.}
      \label{fig:pre_a}
  \end{subfigure}
  \hfill
    \begin{subfigure}[t]{1\textwidth}
      \centering
      \includegraphics[width=0.8\textwidth]{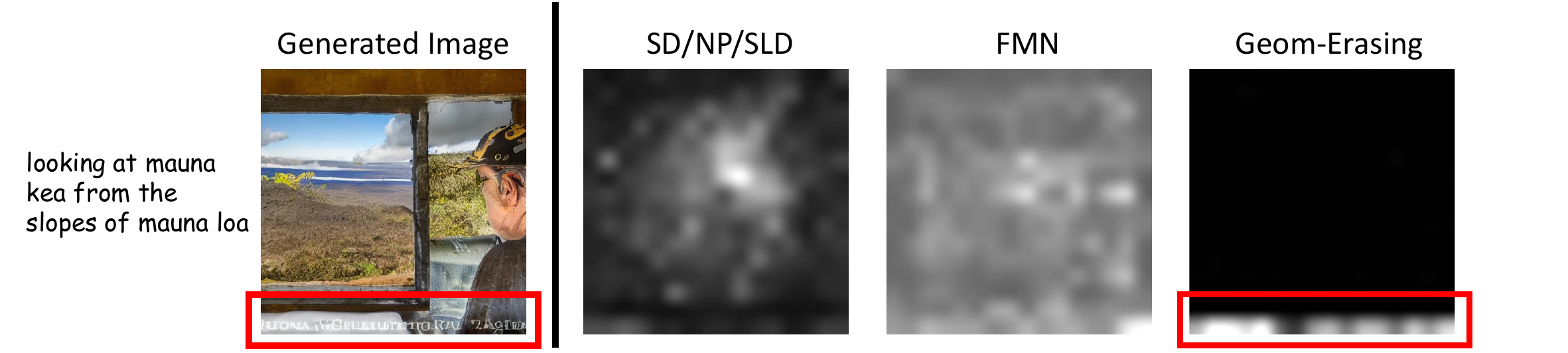}
      \caption{Cross-attention map between keyword "watermark" and the generated image.}
      \label{fig:pre_b}
  \end{subfigure}
  \caption{\textbf{Observations from preliminary experiments.} 
  (a) Implicit concepts cannot be controllably generated. 
(b)  Implicit concepts cannot be recognized by SD.
}
  \label{fig:visual_attn}
  \vspace{-3mm}
\end{figure}

\subsection{Preliminary Experiments}
\label{sec:pre_pre}

In this section, we perform preliminary experiments to examine the limitations of existing approaches~\cite{gandikota2023erasing, kumari2023conceptablation,zhang2023forgetmenot,ho2022classifier,schramowski2023safe}
to erasing implicit concepts. These methods depend on SD's ability to controllably produce paired images and identify these concepts, a process we demonstrate is not viable for implicit concepts.

\subsubsection{Implicit concepts cannot be controllably generated.}
Some erasure methods~\cite{gandikota2023erasing, kumari2023conceptablation} necessitate the creation of image pairs, with one containing the concept for erasure and the other does not.
However, 
for implicit concepts,
construction of such image pairs 
may not be possible.
Our study, depicted in Fig.~\ref{fig:pre_a}, 
reveals that there is a minimal connection 
between \textit{directly asking for ``watermarks'' in the prompt} and \textit{their actual presence in the generated images}. This is supported by a 
Correlation Coefficient 
of $r=-0.08$
and a P-value $p = 0.21$ across 1000
samples, indicating an insignificant effect. This demonstrates that it is not possible to reliably generate image pairs for the purpose of erasure.

\subsubsection{Implicit concepts 
cannot be recognized by SD.} 
Other erasure methods~\cite{zhang2023forgetmenot,ho2022classifier,schramowski2023safe}
depend on the model's ability to identify concepts to be erased. 
However, 
SD often fails to detect the presence of implicit concepts.
Using an example on the implicit concept of watermarks, we visualize the cross-attention map of the last transformer layer between the keyword ``watermark'' and the generated image in Fig.~\ref{fig:pre_b}. 
Since SD~\cite{rombach2022high}, NP~\cite{ho2022classifier} and SLD~\cite{schramowski2023safe} exhibit similar patterns, we visualize them in the same image for brevity.
As can be seen,
SD~\cite{rombach2022high}, NP~\cite{ho2022classifier}, SLD~\cite{schramowski2023safe}, and FMN~\cite{zhang2023forgetmenot} cannot attend to the location of the watermark. 
A more thorough evaluation will be presented in Sec.~\ref{sec:exp_sota}.
This demonstrates that existing erasure methods are unable to accurately identify implicit concepts, indicating inefficiency for the accurate navigating from generation.

\section{Proposed Method}
\label{sec:method}

In this section, we present \methodname, which aims at mitigating the impact of undesired implicit concepts within images and disentangling these concepts from the model. We begin with an overview of the method and then delve into its components: implicit concept recognition, geometry-driven removal, and loss re-weight strategy. 
\begin{figure*}[t]
    \centering
    \includegraphics[width=1\linewidth]{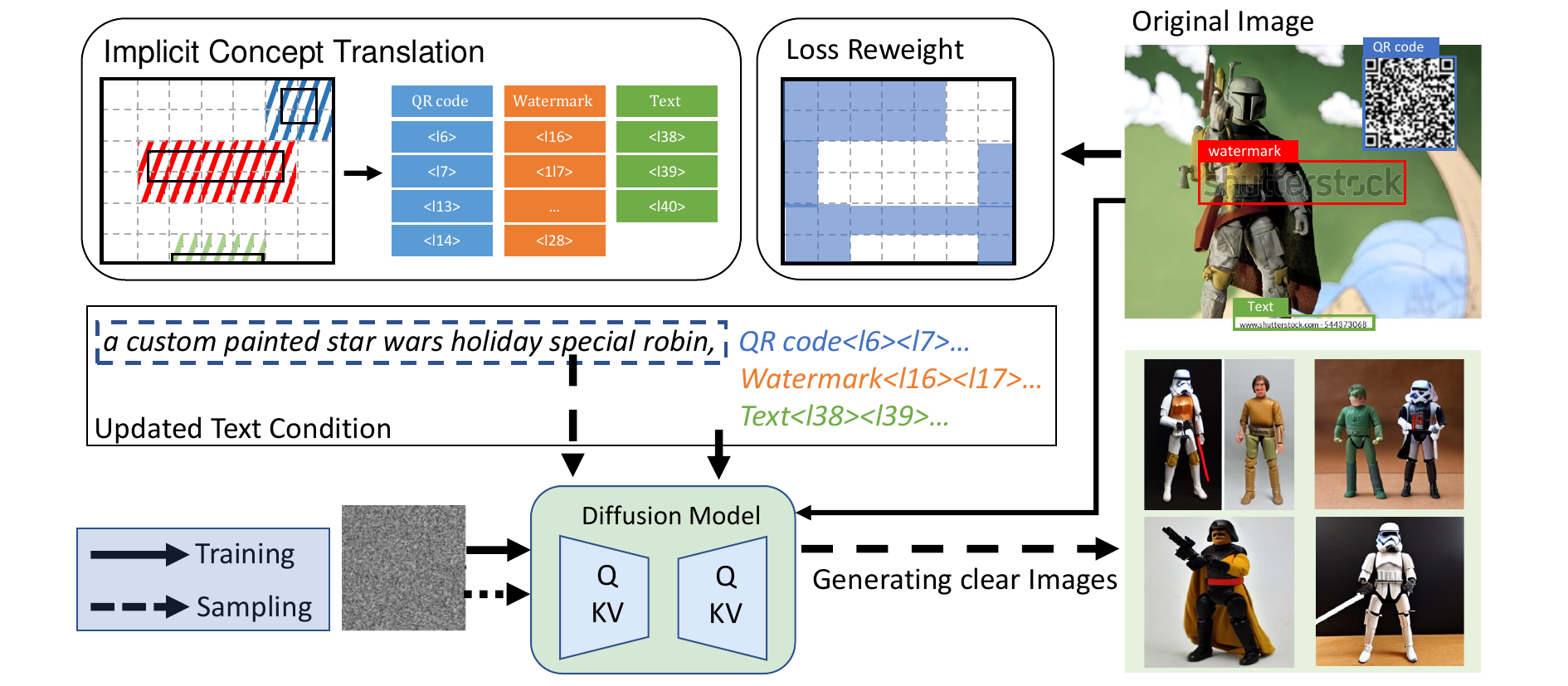}
    \caption{
    \textbf{Model architecture of \methodname.}
    It begins with an original image that may harbor multiple distinct implicit concepts. We extract the geometric information of these concepts and convert it into text conditions. 
    \textbf{Special location tokens} are added to the original text vocabulary representing the bins discretized from the original images.  
    Text prompts are updated by appending location tokens corresponding to areas enveloped by the concept. 
    \textbf{Loss re-weighting} is employed to concentrate more on areas devoid of implicit concepts. 
    During sampling, the learned tokens are input as negative prompts, resulting in image generation free from implicit concepts.
    }
    \label{fig:arch}
\end{figure*}

\subsubsection{Overall architecture.}
The architecture of our method is shown in Fig.~\ref{fig:arch}. The image $x \in X$ may contain various implicit concepts.
Upon confirming the presence of such a concept, we amend the original text condition by appending the concept name
(\eg, \textit{QR codes}, \textit{watermark}, \textit{text} and \textit{unsafeness}). 
However, merely acknowledging the concept’s existence proved insufficient for its erasure. Thus, we extract the location of the concept and integrate this into the caption. 
For \textbf{Model Removal}
setting, optimizing the added tokens representing the existence and location is enough. For \textbf{Data Removal},
we train the diffusion model parameters additionally.
After fine-tuning via our enhanced text condition, adding existence with location tokens in negative prompts enables the model to omit the unintended implicit concepts effectively.

\noindent\textbf{Implicit concept recognition.} The initial step in \methodname involves identifying the presence and location of implicit concepts. Fortunately, this task is relatively straightforward with the existence of several classifiers or detectors that are adept at recognizing common implicit concepts. 
For example, to identify watermarks, 
one can use LAION\footnote{https://github.com/LAION-AI/LAION-5B-WatermarkDetection}.
Details and models for detecting other implicit concepts
are in Appendix~\ref{sec:dataset_details}.

Given a detector
for the implicit concept, we acquire at most $N$ predictions
of the concepts, $L = [p_{i}, (o_i)]^{N}_{i=1}$, where $p_i$ is the confidence in identifying the location as the implicit concept, and $o_i=[a_{i}^{1},b_{i}^{1}, a_{i}^{2},b_{i}^{2}]$ are the coordinates of the concept’s position, where $(a_{i}^{1},b_{i}^{1})$ and $(a_{i}^{2},b_{i}^{2})$ are the upper-left and bottom-right positions. This will be integrated
to our subsequent geometry-driven removal.

\subsubsection{Geometry-driven removal.}
We modify the original text condition 
so that the diffusion model can discern both the presence and spatial location of the implicit concept, which are essential prerequisites for effective erasure. 
Let the original image-text pair from the fine-tuning dataset be $(x,y) \in \mathcal{D}$. If an image is classified as containing a specific concept, we append the concept name to the original text condition:
\begin{align}
    y'= 
\begin{cases}
    y & p_i < t \\
    y \oplus y_{im} & \text{otherwise} \\
\end{cases},
\label{equ:concept_name}
\end{align}
where $t$ is a threshold, $\oplus$ denotes the concatenation operation, and $y_{im}$ is the name of the implicit concept. This enhances the model 
to identify the existence of implicit concept.

To acquaint the model with the geometric information of the concept, we first
discretize the continuous coordinates into bins. Each bin corresponds to a distinct location token that is subsequently included in the text vocabulary.  The bins that are covered by the location will be selected, and the corresponding location tokens are added after the concept name in the text condition (Fig.~\ref{fig:arch}). Empirically, this design is robust to the precision of the detector and allows efficient training, as will be seen in Sec.~\ref{sec:exp}.

Specifically, assume that the image size is $W \times H$, and a bin size of $W_{\text{bin}} \times H_{\text{bin}}$, location tokens are inserted 
into the text vocabulary
$ \langle l \{m,n\} \rangle^{m=W/W_{\text{bin}},n=H/H_{\text{bin}}}_{m=1,n=1}$.
For each implicit concept in the image, the text condition is then updated as:
\begin{equation}
     y' = 
\begin{cases}
    y & \text{if } p_i < t \\
    y \oplus y_{\text{im}} \oplus \langle l \{m,n\} \rangle^{m=A_{\text{bin}}^2,n=B_{\text{bin}}^2}_{m=A_{\text{bin}}^1,n=B_{\text{bin}}^1} & \text{otherwise}
\end{cases}, 
\label{equ:geo}
\end{equation}
where $A_{\text{bin}}^1 = \lfloor a_i^1 / W_{\text{bin}} \rfloor$, $B_{\text{bin}}^1 = \lfloor b_i^1 / H_{\text{bin}} \rfloor$, $A_{\text{bin}}^2 = \lceil a_i^2 / W_{\text{bin}} \rceil$, and $B_{\text{bin}}^2 = \lceil b_i^2 / H_{\text{bin}} \rceil$. This approach 
represents the spatial attributes of implicit concepts within images.

\subsubsection{Loss re-weighting on specific regions.}
Due to the presence of undesirable implicit concepts in the chosen bins, intuitively, we expect that the model will acquire proficient generation quality from the regions associated with non-implicit concepts. Consequently, we assign a fixed lower weight to the loss in order to de-emphasize the implicit concept areas. The resulting refined loss, which incorporates the previously generated bin map, is:
\begin{align}
\label{equ:geo-loss}
     \mathcal{L}_{\text{\methodname}}=\quad \mathbb{E}_{z\sim \mathcal{E}(x), y\sim Y, \epsilon \sim \mathcal{N}(0,1), t} \left[ w \odot \left \| \epsilon - \epsilon_{\theta}(z_t, t, c_\theta(y')) \right \|_2^2 \right], 
\end{align}
where
$\odot$ denotes element-wise multiplication, and
\begin{equation}
w_{m,n} =
    \begin{cases}
      \frac{T}{K + \alpha(T - K)}, &  
          \text{if} \mkern9mu A_{\text{bin}}^1 < m < A_{\text{bin}}^2 
          \text{ and } B_{\text{bin}}^1 < n < B_{\text{bin}}^2 \\
      \frac{\alpha T}{K + \alpha(T - K)}, & \text{otherwise}
    \end{cases}
    \label{equ:w_mn}
\end{equation}
where 
$K=(A_{\text{bin}}^2-A_{\text{bin}}^1) \cdot (B_{\text{bin}}^2 - B_{\text{bin}}^1)$, 
$T$ is the number of bins with
$T = \frac{W}{W_{\text{bin}}} \cdot \frac{H}{H_{\text{bin}}}$, and $\sum w_{m,n} = T$, and
$\alpha$ is a hyperparameter. 
Equation~(\ref{equ:w_mn}) normalizes the weight $w$, aligning the magnitude of Eq.~(\ref{equ:geo-loss}) with that of the original loss. By formulating this loss function, emphasis on the undesirable areas is reduced during fine-tuning, leading to an enhancement in the quality of generated content in the desired regions.

\vspace{-3mm}
\subsubsection{Accessibility of the additional classifier or detector.}
Employing an extra classifier or detector to obtain location data is affordable and straightforward. Numerous models capable of identifying the presence or pinpointing the location of various concepts are readily available, as detailed in Appendix~\ref{sec:dataset_details}. Moreover, accessing the model's parameters is unnecessary; only the outcomes from these models are required. Empirically, 
the effectiveness 
of \methodname 
does not depend heavily on the detector's precision, allowing for some leniency regarding the accuracy of the detector, as will be seen later in Fig.~\ref{fig:geo_acc}.

\section{Experiments}
\label{sec:exp}

In this section,
we perform experiments to demonstrate the effectiveness of \methodname. We first introduce the experimental setup, and then we compare \methodname with several existing erasure methods, followed by the ablation studies on essential components of \methodname.

\subsection{Setup}
\label{sec:setup}

\begin{table}[t]
    \centering
    \setlength\tabcolsep{2pt}
    \scalebox{0.95}{
  \resizebox{1\linewidth}{!}{
    \begin{tabular}{lccccc}
    \toprule
    Dataset Name &  Sample size & ICR & Style & Resolution & Source \\
   \midrule
    ICD-QR & 833 & 25\%    & Cartoon & $512^2$ & Pokemon~\cite{pinkney2022pokemon} \\
    ICD-Watermark & 160k   & 50\%    & Real & $256^2$ & CC12M~\cite{changpinyo2021cc12m} \\
    ICD-Text & 1000k & 100\%   & Real & $256^2$ & LAION~\cite{schuhmann2021laion} \\
    \bottomrule
    \end{tabular}}}%
     \caption{\textbf{Details of Implicit Concept Dataset.} Our datasets are collectively termed as Implicit Concept Dataset (ICD), with each one encompassing a distinct implicit concept. They exhibit variations in several attributes. 
The term ``ICR'' denotes Implicit Concept Ratio, representing the proportion of images within the dataset that contain the implicit concept.}
\vspace{-3mm}
  \label{tab:dataset}
\end{table}%

\textbf{Implicit Concept Dataset.}
We curate three datasets, with 
each one corresponding to an implicit concept. 
The variances in these datasets, in terms of concept types, sizes, Implicit Concept Ratios (ICR), and image styles, are detailed in Table~\ref{tab:dataset}.  In ICD-QR, QR codes are manually embedded in 25\% of the images. ICD-Watermark amalgamates images, with 50\% containing watermarks, sourced from CC12M~\cite{changpinyo2021cc12m}. ICD-Text utilizes a dataset from \cite{yang2023glyphcontrol}, resulting in 100\% of the training images incorporating text. Additionally, corresponding test datasets \textbf{devoid of any implicit concepts} have been assembled for each of the above, to ensure a comprehensive evaluation. More details can be found in Appendix~\ref{sec:dataset_details}.

\subsubsection{Settings.} 
To mirror real-world scenarios, we validate \methodname in two settings: Model Removal and Data Removal. 
For \textbf{Model Removal}, we eliminate the implicit concepts of watermark and unsafeness in the original SD. Watermark is evaluated under the evaluation set of our constructed dataset ICD-watermark, and unsafeness is evaluated under the I2P benchmark. 
\textbf{Data Removal} mimics personalized fine-tuning when practicers collect downstream datasets that may also contain different implicit concepts, attributed to the limitations in sources and collecting methods. This setting is evaluated under all three ICD datasets.

\begin{table}[t]
\setlength\tabcolsep{2pt}
  \centering
   \resizebox{1\linewidth}{!}{
    \begin{tabular}{l|ccc|cccccccc|c}
    \toprule
      & \multicolumn{3}{c|}{Watermark}& \multicolumn{8}{c|}{Unsafeness (I2P benchmark~\cite{schramowski2023safe})} & Expected Max. \\
      & FID & ICR & $F*R/100$ & Hate & Harassment & Violence & Self-harm & Sexual & Shocking & Illegal activities & Overall & Inappro. \\
    \midrule
    SD~\cite{rombach2022high}    & 9.05  & 11.13  & 1.01 & 0.40   & 0.34  & 0.40   & 0.40   & 0.30   & 0.51  & 0.36  & 0.39  & $0.97_{0.06}$ \\
    ESD~\cite{gandikota2023erasing}   & 9.49  & 11.28  & 1.07 & 0.17  & 0.16  & 0.24  & 0.22  & 0.17  & 0.16  & 0.22  & 0.19  &  -\\
    FMN~\cite{zhang2023forgetmenot}   & 10.05  & 10.83  & 1.09 &  -     &   -    &    -   &   -   &   -    &  -     &  -     &  -     & - \\
    NP~\cite{ho2022classifier}    & 9.12  & 11.13  & 1.02 & 0.16  & 0.14  & 0.19  & 0.14  & 0.08  & 0.25  & 0.13  & 0.16  & $0.80_{0.18}$ \\
    SLD-Strong~\cite{schramowski2023safe}   & 9.87  & 9.92  & 0.98 & 0.15  & 0.13  & 0.17  & 0.19  & 0.09  & 0.20   & 0.09  & 0.13  & $0.72_{0.19}$ \\
    \midrule
    \rowcolor{backcolor}
    \small{\methodname} & \textbf{8.34}  & \textbf{7.31}  & \textbf{0.61} & \textbf{0.11}  & \textbf{0.11}  & \textbf{0.13}  & \textbf{0.06}  & \textbf{0.05}  & \textbf{0.15}  & \textbf{0.07} & \textbf{0.09}  & $\textbf{0.63}_{\textbf{0.20}}$ \\
    \bottomrule
    \end{tabular}}%
    \caption{\textbf{Comparison between \methodname and existing erasure methods under Model Removal setting.} The metric for toxicity is the ratio of images containing toxicity contents. The last column is the bootstrap estimates of a model containing toxic images at least once for 25 prompts~\cite{schramowski2023safe}. \methodname achieves the state-of-the-art when eliminating these two implicit concepts.}
    \vspace{-3mm}
  \label{tab:sota_setting1}%
\end{table}%

\begin{table}[t]
  \centering
  \setlength\tabcolsep{8pt}
  \scalebox{1}{
  \resizebox{1\linewidth}{!}{
  \begin{tabular}{l|ccc|ccc|ccc}
    \toprule
          & \multicolumn{3}{c|}{ICD-QR} & \multicolumn{3}{c|}{ICD-Watermark} & \multicolumn{3}{c}{ICD-Text} \\
          & FID & ICR & $F*R/100$ & FID & ICR & $F*R/100$ & FID & ICR & $F*R/100$ \\
    \midrule
   SD~\cite{rombach2022high}    & 65.82  & 74.59  & 49.10  & 7.59  & 30.40  & 2.31  & 54.23  & 71.84  & 38.96  \\
    ESD~\cite{gandikota2023erasing}   & 90.97  & 17.64  & 16.05  & 7.64  & 28.98  & 2.21  & 60.56  & 38.08  & 23.06  \\
    FMN~\cite{zhang2023forgetmenot}   & 71.76  & 80.42  & 57.71  & 7.79  & 30.76  & 2.40  & 57.38  & 74.75  & 42.89  \\
    NP~\cite{ho2022classifier}    & 69.31  & 59.64  & 41.34  & 7.54  & 27.71  & 2.09  & 52.13  & 65.63  & 34.21  \\
    SLD-Strong~\cite{schramowski2023safe}   & 80.05  & 70.25  & 56.24  & 8.56  & 32.56  & 2.79  & 55.36  & 66.08  & 36.58  \\
    \midrule
    \rowcolor{backcolor}
    \methodname  & \textbf{41.41}  & \textbf{5.38}  & \textbf{2.23}  & \textbf{6.99}  & \textbf{5.02}  & \textbf{0.35}  & \textbf{38.74}  & \textbf{13.48}  & \textbf{5.22}  \\
    \bottomrule
    \end{tabular}}}
      \caption{\textbf{Comparison between \methodname and other erasure methods under Data Removal setting.} All models are evaluated under ICD.
  \methodname achieves the best among all three criteria, showing the successful elimination of the implicit concept while improving the image quality.}
  \label{tab:sota}%
  \vspace{-2em}
\end{table}%

\vspace{-3mm}
\subsubsection{Baselines and evaluation metrics.}
We compare \methodname with existing erasure methods, including Erased Stable Diffusion (ESD)~\cite{gandikota2023erasing}, Forget-Me-Not (FMN)~\cite{zhang2023forgetmenot}, Negative Prompt (NP)~\cite{ho2022classifier}, and Safe Latent Diffusion (SLD)~\cite{schramowski2023safe}. 
To depict the outcomes, we employ the Frechet Inception Distance (FID)~\cite{heusel2017gans} and the Implicit Concept Ratio (ICR) which is defined as\textit{ the ratio of the number of images containing implicit concept to the total number of images}. 
\textbf{Both metrics prefer lower values.}
To facilitate a more comprehensive comparison between models and offer an integrated perspective on performance concerning both metrics, we introduce the $F*R/100$, which is calculated as the product of FID and ICR, serving as a unified metric for evaluating model performance.
We also adopt the Inappropriate Image Prompts (I2P) benchmark~\cite{schramowski2023safe} to validate our erasure of toxicity in the pre-trained diffusion model. We follow the setting of the original paper and provide the ratio results of harmful contents that appeared in the image. Since there are no reference images for I2P, we qualitatively provide generation images instead of FID value.

\subsection{Comparison with previous methods}
\label{sec:exp_sota}

\begin{figure*}[t]
    \centering
    \includegraphics[width=1\linewidth]{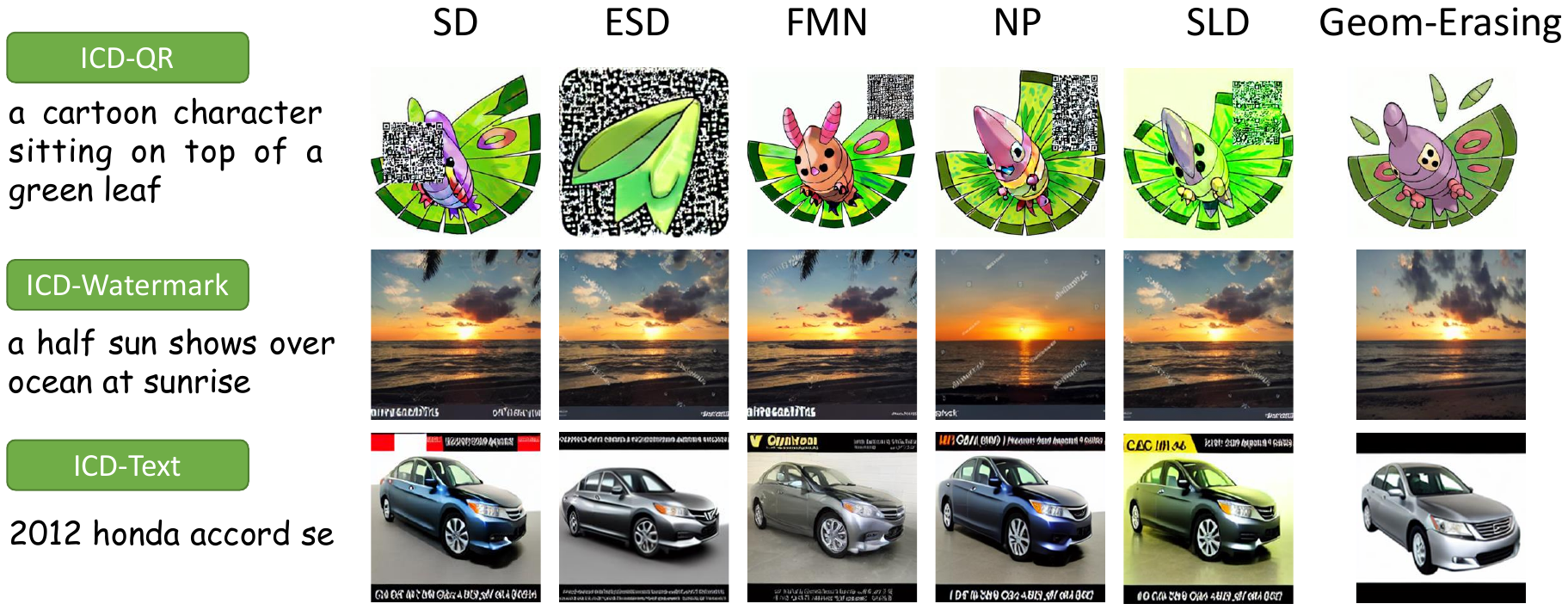}
    \caption{
    \textbf{Qualitative comparison between different methods.} Existing methods cannot erase ICs effectively while \methodname successfully avoids the generation.
    Check more comparison in Fig.~\ref{fig:erase_sd} and~\ref{fig:sota_more}.
    }
    \vspace{-3mm}
    \label{fig:comp_w_baseline}
\end{figure*}

\subsubsection{Erasure of pre-trained SD (model removal).}
We first compare \methodname with other methods in the setting of erasing ``watermark''and ``unsafeness'' contained in the pre-trained SD. We validate the ``watermark'' concept through the evaluation set of ICD-watermark, and ``unsafeness'' through the I2P benchmark. As shown in Table~\ref{tab:sota_setting1}, \methodname performs the best in both implicit concepts, achieving new state-of-the-art results on the I2P benchmark.

\vspace{-3mm}
\subsubsection{Erasure in personalized fine-tuning (data removal).} 
As illustrated in Table~\ref{tab:sota} and visualized in Fig.~\ref{fig:comp_w_baseline}, \methodname notably surpasses existing erasure methods across three distinct implicit concepts by substantially diminishing their occurrence in the synthesized images. Even in instances where fine-tuned images all contained text, our method remarkably reduces text presence to just \textbf{13.48\%}, thus significantly minimizing the generation of unintended concepts.

\vspace{-3mm}
\subsubsection{Removing implicit concepts improves generation.} Besides reducing unintended elements, \methodname also improves the generation quality compared to other methods in both settings. This improvement is due to the method’s ability to effectively erase implicit concepts—since the ideal images (reference images) don’t contain these elements, avoiding them results in higher quality scores (FID scores). Further insights into the correlation between erasure efficacy and enhanced image quality are detailed in our ablation study in Sec.~\ref{sec:ablation}. 

\vspace{-3mm}
\subsubsection{Existing erasure methods cannot do well for implicit concepts.}
The methods of FMN~\cite{zhang2023forgetmenot}, NP~\cite{ho2022classifier}, and SLD~\cite{schramowski2023safe} demonstrate limitations in effectively removing implicit concepts. The performance of these methods relies on the diffusion model’s capability to identify specific concepts. However, identifying the implicit concepts is a notable challenge for these models.
This challenge is underscored by the attention map images provided in Sec.~\ref{sec:pre_pre}, which depicts the models' inadequacies in accurately identifying and addressing the implicit concepts, subsequently hindering successful erasure. 
Among all the baselines, we find ESD \cite{gandikota2023erasing} demonstrates superior erasure performance, albeit with a higher FID score. This can be attributed to the approach employed by ESD, where the fine-tuned SD model is trained to move away from the images it originally generated, regardless of whether they contain the intended concept or not. However, since original generated images may contain implicit concepts with high probability, ESD might result in unintended concept removal while affecting meaningful one, as in the 2nd column of Fig.~\ref{fig:sota_more}.

In contrast, our proposed method, \methodname, demonstrates the ability to effectively remove implicit concepts while preserving the other meaningful concepts, yielding favorable ICR and FID results. It surpasses the state-of-the-art, as evidenced by the superior $F*R/100$ measure.
Refer to Appendix~\ref{sec:toxicity_visual} for visual comparisons among different methods. \methodname offers a more refined and precise erasure process, ensuring that only the targeted implicit concept is removed, without affecting other relevant concepts.

\begin{figure}[t]
    \begin{minipage}[b]{0.58\linewidth}
    \centering
    \resizebox{\linewidth}{!}{
    \begin{tabular}{ccc|ccc}
    \toprule
     Concept & \quad Geometric & Loss re-weight & FID & ICR & $F*R$ \\
    \midrule
          &       &       & 7.59  & 30.40  & 230.74  \\
    \midrule
    $\surd$   &       &       & 7.06  & 17.04 & 120.30  \\
          & $\surd$   &       & 6.97  & 11.18  & 77.92 \\
          &       & $\surd$   & 6.46 &  29.38  &  189.79 \\
    \midrule
    $\surd$   & $\surd$   &       & 6.81  & 7.36  & 50.12  \\
    \rowcolor{Gray}$\surd$   & $\surd$   & $\surd$   & \textbf{6.42} &	7.23 &	\textbf{46.42} \\
    \midrule
    \multicolumn{3}{l|}{\textit{Fine-tuning with 0\% watermark (oracle)}} & 6.93 & \textbf{7.13} & 49.41 \\
    \bottomrule
    \end{tabular}}
    \captionof{table}{\textbf{Ablation of different components.}
  Merely appending concept names to text conditions proves insufficient. Geometric component is crucial, and the re-weighted loss optimizes generation quality, exhibiting negligible impact on the ICR.
  Default settings are marked in \colorbox{Gray}{gray}.
  }
  \vspace{-3mm}
  \label{tab:ablative}
    \end{minipage}%
    \hfill
\begin{minipage}[b]{0.4\textwidth}
\centering
    \includegraphics[width=1.05\textwidth]{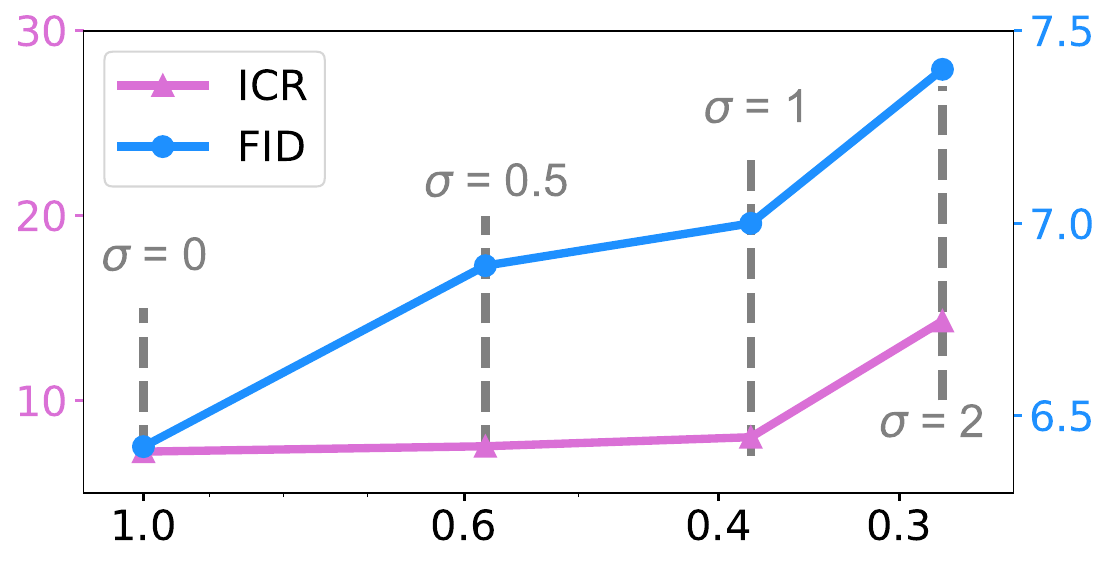}
    \caption{\textbf{Effects of geometric accuracy.} 
    The x-axis indicates the 
 IoU decided by the noise $\sigma$ added on the location. \methodname has a tolerance of around 0.4 IoU considering ICR.}
    \vspace{-3mm}
    \label{fig:geo_acc}
\end{minipage}
\end{figure}

\subsection{Ablation and Analysis}
\label{sec:ablation}
In this section, we perform various ablations analyses to demonstrate the effectiveness of different components in \methodname. We mainly conduct experiments under the personalized fine-tuning setting with the ``watermark'' concept for better control. Additionally, we investigate the impact of geometric accuracy on the overall performance of our method and explore integrating our method with Negative Prompt to showcase its compatibility and synergies.

\begin{figure*}[t]
    \centering
    \includegraphics[width=1\linewidth]{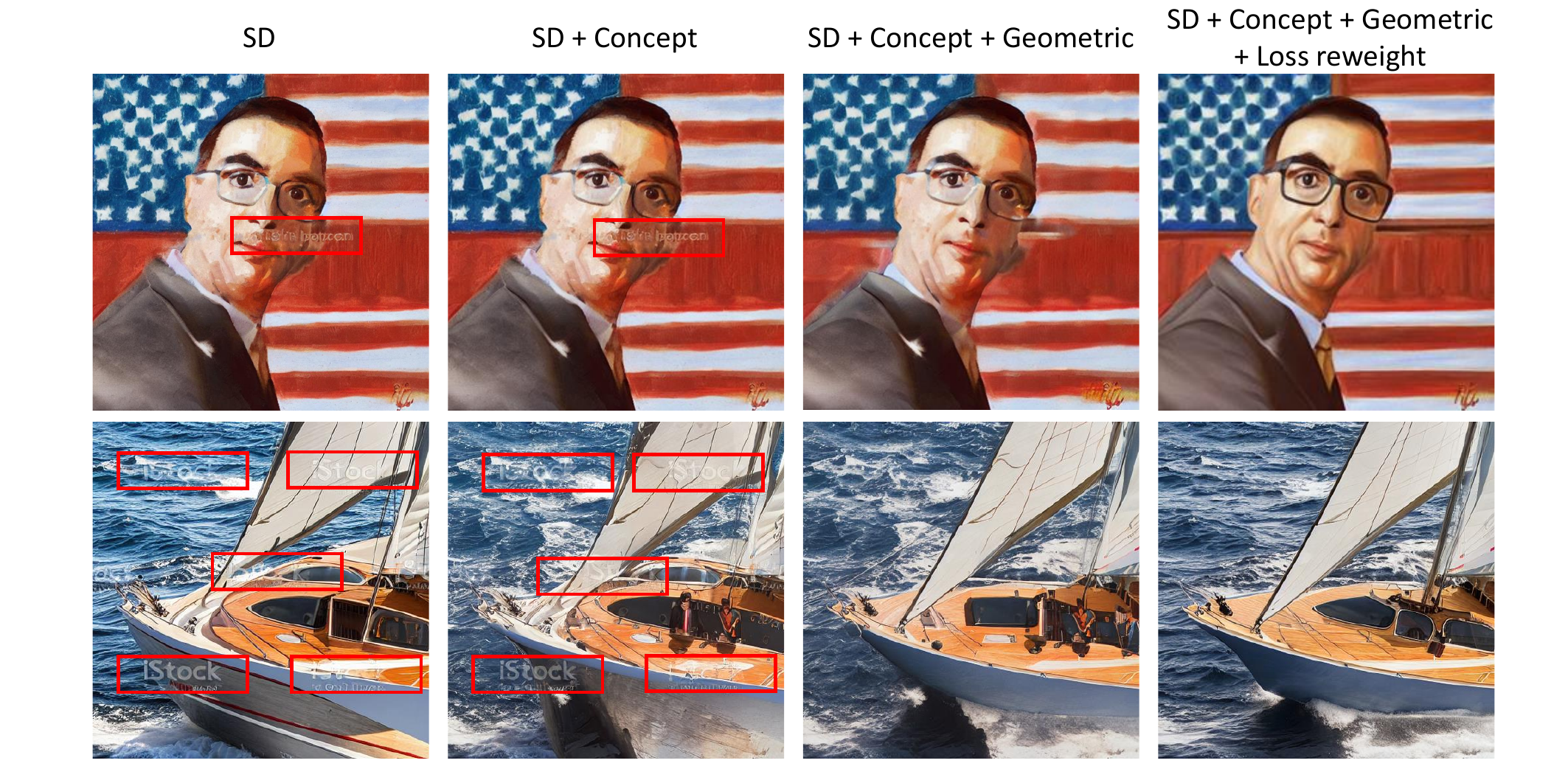}
    \caption{
    \textbf{Visualization of different components.} Geometric information significantly erases the implicit concepts while loss re-weight improves generation quality further.
    }
    \label{fig:ablation}
\end{figure*}

\vspace{-3mm}
\subsubsection{Ablative analysis.}
The ablation results, as shown in Table~\ref{tab:ablative} and Fig.~\ref{fig:ablation}, shed light on the importance of different components in our methods. We separately integrate the concept name (Eq.~\ref{equ:concept_name}), geometric information (Eq.~\ref{equ:geo}), and loss re-weight (Eq.~\ref{equ:geo-loss}) before combining them. Notably, the geometric component proved pivotal, markedly reducing both FID and ICR, particularly when synergized with the concept condition, enhancing the model’s overall performance.
The loss re-weight component contributes to improving the visual appeal of the generated images while maintaining the efficacy of implicit concept removal. Throughout the ablative studies, a consistent trend between FID and ICR is observed, implying enhanced erasure correlates to superior image quality.
Moreover, when fine-tuning the model with no implicit concepts (0\% images with watermarks), the model, in its optimal state, achieves an ICR comparable to \methodname. Interestingly, \methodname surpasses even this optimum in FID and $F*R$, emphasizing the importance of geometric information in refining concept learning and subsequently improving image quality.

\subsubsection{Choice of bin size, number of selected bins, and re-weight loss} influence erasure outcomes. Table~\ref{tab:cam_size_k} depicts the effects of varying bin sizes $M$ and selected bin numbers $K$, with bold values denoting optimal performance and the gray row signifying the default selection. Bins are ranked and selected by value $p_i$ as stated in Sec.~\ref{sec:method}. Initially, we analyze the impact of bin sizes by fixing the ratio between the number of selected bins and bin size and varying the size from $8^2$ to $64^2$. As the bin size increases, the performance initially improves and then starts to decline. This trend suggests that higher resolutions may provide more accurate concept localization but can also dilute the information density of the original text. Subsequently, with a fixed bin size, varying the number of selected bins showed enhancement in erasure performance up to a saturation point.

For re-weight loss, we conduct ablation experiments based on the model in the gray row of Table~\ref{tab:cam_size_k}. An alternate re-weight loss incorporating the $p_i$ values is proposed. As shown in Table~\ref{tab:reweight_loss}, applying the re-weight loss leads to improved FID compared to the model without the loss. However, 
utilizing the $p_i$ may degrade the erasure performance. Opting for simplicity and effectiveness, a fixed value is utilized for the area covered by implicit concept, as outlined in Eq.~\ref{equ:w_mn}.

\subsubsection{Geometric accuracy.}
We execute experiments to investigate the sensitivity of \methodname to the precision of geometric information provided. Upon selecting the bins, we introduce two noise scalars, $\epsilon_1, \epsilon_2 \sim \mathcal{N}(0,\sigma^2)$, to the selected index, illustrated as $y' = y \oplus y_{\text{im}} \oplus \langle l \{m+\epsilon_1,n+\epsilon_2\} \rangle$. Variations in $\sigma$ yield distinct IoU values between the originally selected and noised bins, visualized in Fig.~\ref{fig:geo_acc}. The ICR can tolerate a geometry accuracy up to 0.4 IoU; however, the erasure performance experiences a decline as accuracy continues to decrease.

\vspace{-3mm}
\subsubsection{Negative prompt.}
We aim to improve our method by adding both the learned concept name and geometric information to the negative prompt to better erase unwanted details. Since we want implicit concepts not to appear in any part of the image, we use location tokens that are picked uniformly or randomly as negative prompts. As shown in Table~\ref{tab:np}, adding the concept name to the negative prompt improves the erasure and overall quality of generated content due to the model's improved ability to recognize concepts. Adding geometric information, whether uniformly or randomly, further improves the erasure, but it also tends to increase the FID. We plan to explore the reasons for this increase in more detail in future work.

\begin{table*}[t]
\begin{minipage}{0.3\columnwidth}
    \centering
     \resizebox{0.928\linewidth}{!}{
   \begin{tabular}{cc|ccc}
    \toprule
    \multirow{2}{*}{M} &\multirow{2}{*}{K} &
    \multirow{2}{*}{FID}   & 
    \multirow{2}{*}{ICR}   & 
    \multirow{2}{*}{$F*R$}   \\
    &  & & & \\
    \midrule
    $8^2$  & 4     & 6.78  & 18.26 & 123.80  \\
    $16^2$ & 16    & 6.53  & 15.45 & 100.89  \\
    $32^2$ & 64    & \textbf{6.51}  & 12.64 & 82.29  \\
    $64^2$ & 150   & 7.29  & 16.47 & 120.07  \\
    \rowcolor{Gray}
    $32^2$ & 72 & 6.81  & \textbf{7.36} & \textbf{50.12}  \\
    $32^2$ & 80    & 6.92  & 7.35  & 50.86  \\
    \bottomrule
    \end{tabular}}
    \caption{\textbf{Bin size and selected bins.} Larger bin size and more bins improve results.}
    \vspace{-6mm}
    \label{tab:cam_size_k}
\end{minipage}\hfill
\begin{minipage}{0.3\columnwidth}
    \centering
     \resizebox{1\linewidth}{!}{
    \begin{tabular}{cc|ccc}
    \toprule
    Re-weight & \multirow{2}{*}{$\alpha$} & \multirow{2}{*}{FID}   & \multirow{2}{*}{ICR} & \multirow{2}{*}{$F*R$} \\
    Function & & & & \\
    \midrule
    \rowcolor{Gray}
     Eq.~\ref{equ:w_mn} &  0.25 & 6.42 & \textbf{7.23} & \textbf{46.42 } \\
          & 0.50   & 6.40   & 7.63  & 48.83  \\
          & 0.75  & 6.45  & 7.41  & 47.79  \\
    \midrule
    $(1-p_i)^\alpha$ &  0.5   & 6.27  & 9.21  & 57.75  \\
          & 1 & 6.33 & 9.34 & 46.46  \\
          & 2     & \textbf{6.26}  & 9.44  & 59.09  \\
    \bottomrule
    \end{tabular}}
    \vspace{+2.7mm}
    \caption{\textbf{Re-weight loss.} A fixed re-weight design is good enough for better ICR.}
    \vspace{-6mm}
    \label{tab:reweight_loss}
\end{minipage}\hfill
\begin{minipage}{0.3\columnwidth}
    \centering
     \resizebox{1\linewidth}{!}{
    \begin{tabular}{lccc}
    \toprule
    Negative Prompt & \multicolumn{1}{l}{FID} & \multicolumn{1}{l}{ICR} & \multicolumn{1}{l}{$F*R$} \\
    \midrule
    w/o NP & 6.42  & 7.23  & 46.42 \\
    \midrule
    Concept & \textbf{6.15}  & 7.03  & 43.23 \\
    \rowcolor{Gray}
    + Uniform Geometry &  6.99   & 5.02  & \textbf{35.09} \\
    + Random Geometry &  7.12  & \textbf{4.98} & 35.46 \\
    \bottomrule
    \\
    \\
    \\
    \\
    \\
    \end{tabular}}
    \vspace{+2mm}
    \caption{\textbf{Input as negative prompt.} Usage of the geometry improves ICR further.}
    \vspace{-6mm}
    \label{tab:np}
\end{minipage}
\end{table*}

\section{Conclusion}
\label{sec:conclusion}
Fine-tuning on personalized datasets is a prevalent practice, but the presence of unwanted implicit concepts like QR codes, watermarks, and text within these datasets can pose significant challenges during the refinement of personal diffusion models. This paper delves into the substantial impact of such implicit concepts, establishing a formal framework for their removal. Conventional methods, which predominantly depend on pre-trained diffusion models or merely acknowledge concept existence, falter in eradicating these implicit elements. To address this, we introduce \methodname, a novel approach that incorporates geometric information during the fine-tuning phase, translating this information to the text domain and refining the initial text condition. We substantiate our approach through three diverse datasets, each laden with distinct implicit concepts. The exemplary performance of \methodname underscores its efficacy in eradicating specific concepts, paving the way for enhanced model fine-tuning practices. 

\section*{Acknowledgement}
We gratefully acknowledge the support of MindSpore, CANN (Compute Architecture for Neural Networks) and Ascend AI Processor used for this research.
This work was partially supported by 
the Research Grants
Council of the Hong Kong Special Administrative Region
(Grants  C7004-22G-1 and 16202523).
This research has been made possible by funding support from the Research Grants Council of Hong Kong through the Research Impact Fund project R6003-21.

%
%
\bibliographystyle{splncs04}
\bibliography{main}

\clearpage

\appendix

\setcounter{table}{0}
\setcounter{figure}{0}
\renewcommand{\thetable}{A\arabic{table}}
\renewcommand{\thefigure}{A\arabic{figure}}

\section{Details of Training and Datasets}
\label{sec:dataset_details}

We perform erasure in two distinct settings. For \textbf{Model Removal}, we remove the \textit{watermark} and \textit{unsafeness} concepts from the original Stable Diffusion v1-5, and for \textbf{Data Removal}, we simulate the scenario where users fine-tune diffusion models with personalized datasets containing diverse implicit concepts. Detailed training information and dataset specifics for both settings are presented herein.

\subsection{Model Removal: Erasing from pre-trained SD}
In this setting, we identify text prompts likely to elicit the generation of images with implicit concepts and apply geometric erasure to these identified concepts.

\subsubsection{Watermark Removal.} For \methodname training, 5000 text prompts inducing watermark-containing images are collected from ICD-Watermark. We use the watermark classifier model\footnote{https://github.com/LAION-AI/LAION-5B-WatermarkDetection} to provide the concept existence condition and its classifier activation map for location information. 
The model is trained and only updates the added tokens representing concept existence and location. During generation, we use the original text condition as positive guidance and the existence and location tokens as negative guidance. The results reported in the main paper are evaluated on the test dataset of ICD-Watermark. With such a method, \methodname is able to generate images without a watermark and keep other contents the same as not using learned tokens as negative prompts. See visualizations in the first two columns of Fig.~\ref{fig:erase_sd}.

\subsubsection{Toxicity Removal.} Following the Inappropriate Image Prompts (I2P)~\cite{schramowski2023safe} setting, 47030 images encompassing 7 toxicity categories are generated: hate, harassment, violence, self-harm, sexual content, shocking images, illegal activity. NudeNet\cite{nudenet} and Q-16 classifier~\cite{schramowski2022can} are employed to identify toxicity, considering both existence and location for the ``sexual content'' category using NudeNet, while only considering existence for other categories due to unavailable location information. The model is trained and only updates the added tokens. We apply the same generation strategy as `Watermark Removal'. Visualization can be seen in the last two columns of Fig.~\ref{fig:erase_sd}.

\subsection{Data Removal: Erasing from ICD}
\begin{figure*}[t]
  \centering
\includegraphics[width=0.95\linewidth]{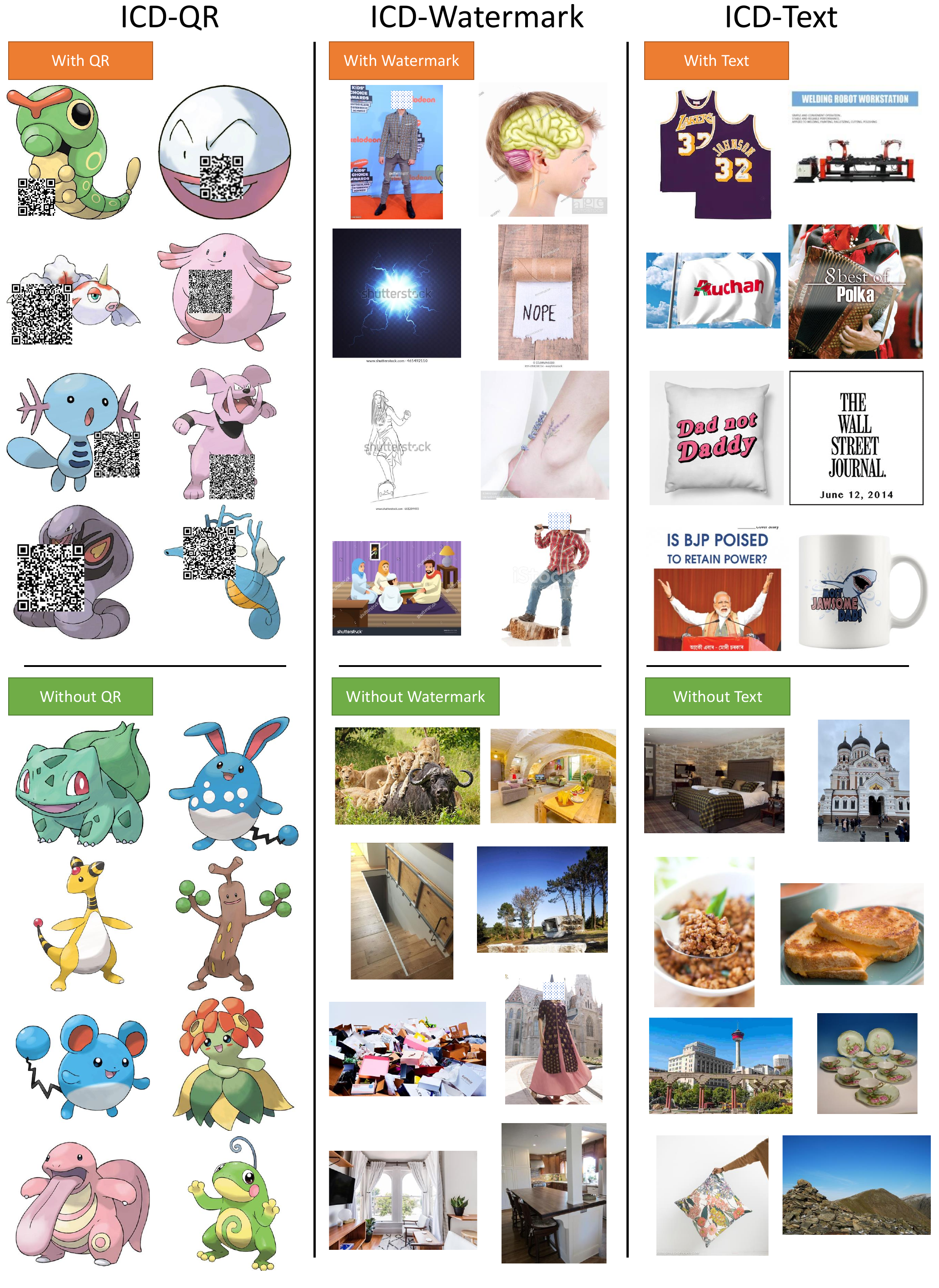}
  \caption{\textbf{Image samples from our Implicit Concept Dataset (ICD).} We provide both images with implicit conceptes}
  \label{fig:dataset_samples}
\end{figure*}

In this scenario, we assemble three datasets (ICD) to emulate user fine-tuning of stable diffusion with personalized datasets harboring diverse implicit concepts. These concepts arise due to constraints in image sources and collection methods within specific downstream tasks. Further details on training and datasets are provided, with samples displayed in Fig.~\ref{fig:dataset_samples}.

\subsubsection{ICD-QR (QR code Removal).} Real QR codes to the Pokemon dataset~\cite{pinkney2022pokemon}. The total dataset contains 802 image-text pairs, which is divided into two portions: 80\% for fine-tuning, and the remaining 20\% for testing. 
In the training subset, QR codes are pasted to 25\% of the images, with QR code lengths varying from 1/4 to 1/2 of the image length, placed randomly, occasionally overlapping with the original content to resemble real-world scenarios. Importantly, test images remain QR code-free for evaluation. To provide concept conditions and geometric information for our method and evaluation, a Faster-RCNN detector is trained using an open-source QR detection dataset\footnote{https://universe.roboflow.com/roboflow-qsmu6/qr-codes-detection}. We use the revised text conditions and fine-tune all the parameters of SD. Generation results can be seen in the first two rows of Fig.~\ref{fig:sota_more}. 

\subsubsection{ICD-Watermark (Watermark Removal).} 
Images are collected from CC12M~\cite{changpinyo2021cc12m}, amounting to 320k images, with half containing watermarks. A watermark recognition tool trained by LAION is employed to identify watermarked images from CC12M with a high confidence threshold of 0.9 to ensure accuracy. For preliminary experiments, subsets of 160k images with varying ratios of watermarked images are constructed. In other experiments, a consistent dataset of 80k images with watermarks and 80k images without watermarks is selected. To provide concept conditions, the watermark recognition tool is used, and for geometric information, the classifier activation map produced by the tool is employed, deciding areas of containing watermarks. Refer to the middle two rows of Fig.~\ref{fig:sota_more} for more generation results.

\vspace{-3mm}
\subsubsection{ICD-Text (Text Removal).} Text images are gathered from LAION~\cite{schuhmann2021laion}. The training dataset we used is provided by \cite{yang2023glyphcontrol}, known as LAION-Glyph. It comprises 1M samples, with each image containing text. For the evaluation dataset, 2k text-free images are collected. To obtain geometric information and for evaluation purposes, PP-OCRv3~\cite{du2021pp} is used to detect text within the images. See the last two rows of Fig.~\ref{fig:sota_more} for visualization.

\section{Why Implicit Concepts exists?}
\label{sec:app_preli}
To explore the reason and difficulty of this problem, we aim to elucidate the concept of watermarking, denoted as $y_{im}$=`watermark'.
We first analyze the reason why implicit concept exists, and then we ascertain the impact of this implicit concept through following experiments.

\subsubsection{Implicit concept in SD stems from the implicit concept in data.} 
\begin{figure}[t]
     \centering
     \begin{subfigure}[b]{0.46\textwidth}
         \centering
         \includegraphics[width=1\linewidth]{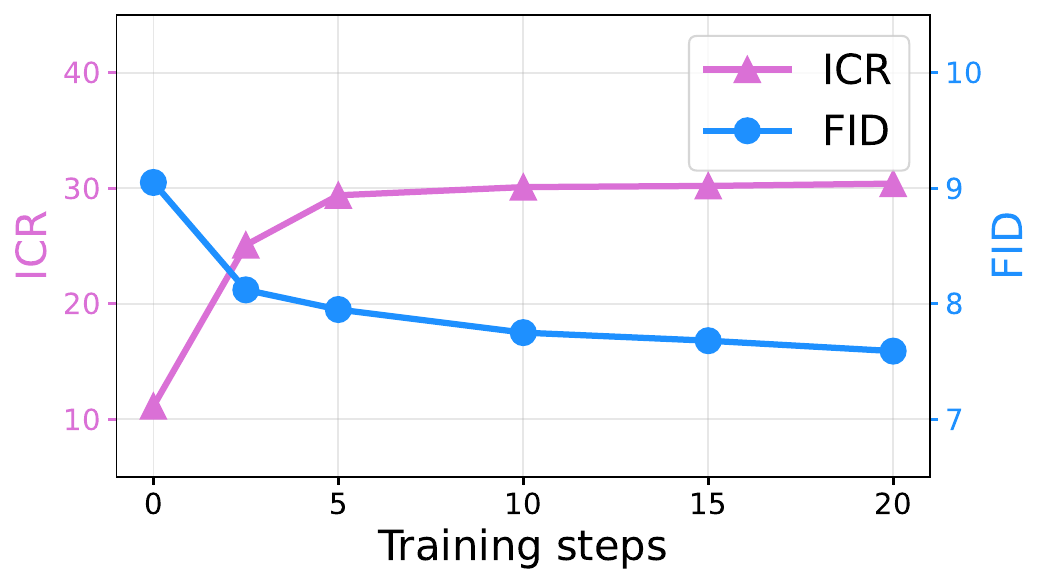}
         \caption{Trade-off between ICR and FID.}
         \label{fig:app_pre_a}
     \end{subfigure}\hfill
     \begin{subfigure}[b]{0.46\textwidth}
         \centering
         \includegraphics[width=1\linewidth]{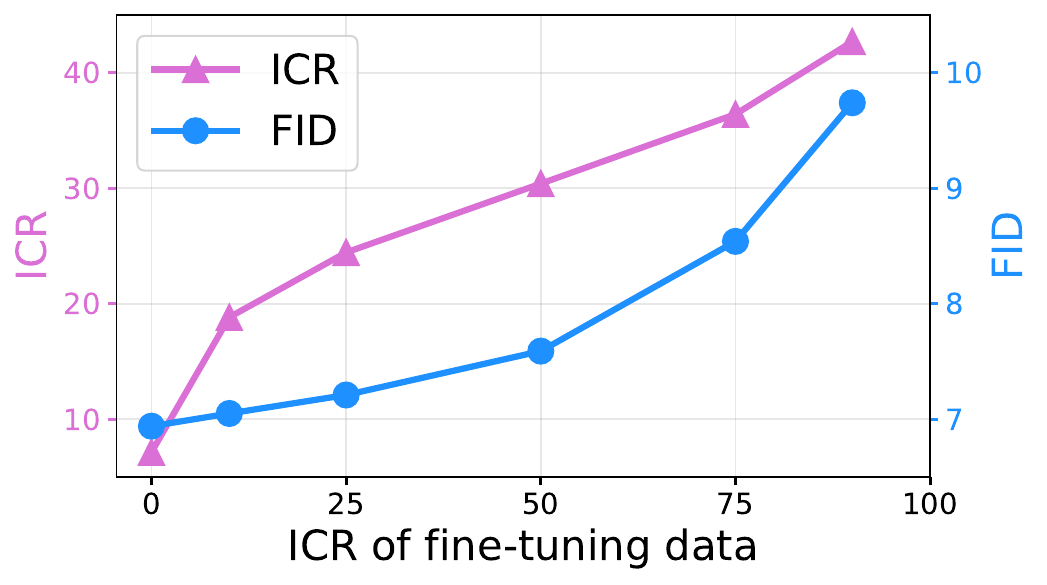}
         \caption{Consistent trend between both.}
         \label{fig:app_pre_b}
     \end{subfigure}
        \caption{\textbf{The severity of implicit concept}. 
   In both figures, the left y-axis represents the ICR, while the right y-axis represents FID.
   \textbf{(a)} When tuning with a 50\% concept ratio, the generated image ratio increases, while FID continues to improve during the tuning phase, suggesting a trade-off between image quality and the presence of implicit concepts. 
   It is worth noting that the original SD model, represented by the 0-th training step, still generates over 10\% watermarked images.
   \textbf{(b)} 
   The ratio of the implicit concept is varied in the fine-tuning dataset. Higher ratios in the fine-tuning data correspond to higher ratios in the generated images, leading to poor image quality. This highlights the severity of the problem related to implicit concept presence.}
   \vspace{-3mm}
    \label{fig:preli}
\end{figure}

The presence of implicit concepts in the Stable Diffusion (SD) model is rooted in implicit concepts present in the training data. We assume that these implicit concepts emerge because, during the training phase, \textit{they exist in the images without specific words in the text conditions expressing them}. 
To understand this, we conducted preliminary studies involving dataset manipulation to explore the causal relationship between training data characteristics and the severity of this issue. 
To emulate real-world scenarios, we curate a fine-tuning dataset, ICD-Watermark, from CC12M. In this dataset, 50\% of the images contain watermarks, and the rest do not, with no ``watermark'' keyword present in text conditions.
The SD model is subsequently fine-tuned with Eq.~\ref{equ:sd_loss} to stimulate the training process. Evaluation is focuses on the implicit concept ratio (ICR, the ratio of images containing implicit concepts, defined in Sec.~\ref{sec:setup}) and the Fréchet Inception Distance (FID) of the synthesized images. 
As can be seen in Fig.~\ref{fig:app_pre_a}, the training step 0 represents the original SD model without any fine-tuning. 
As fine-tuning progresses, FID typically decreases, but the proportion of watermarks in the generated images (ICR) steadily rises, indicating a trade-off between these two metrics. 
This indicates that models trained with datasets containing implicit concepts tend to unconsciously replicate these elements during generation. indicating such concepts are implicitly learned in the model.
This poses a challenge as we aim to achieve a model that closely resembles the target domain while being free of implicit concepts, especially when conducting personalized fine-tuning.

\vspace{-3mm}
\subsubsection{Severity of implicit concepts.}
We further investigate it by examining the impact of varying watermark ratios in fine-tuning datasets.
Multiple datasets are created with a consistent number of total images but different proportions of watermarked images.
The results, depicted in Fig.~\ref{fig:app_pre_b}, reveal a consistent pattern. When the model is fine-tuned with a higher proportion of watermarked images, the generated images also exhibit a higher watermark ratio. Furthermore, this will also affect the generation quality of images as the FID score continues to be worse. 

In summary, our preliminary experiments indicate that training with datasets containing implicit concepts markedly deteriorates the performance and introduces a considerable amount of unwanted concepts, motivating our proposed \methodname that can achieve both high generation quality and free of implicit concepts.

\section{Additional Erasure visualizations}
\label{sec:toxicity_visual}
\begin{figure*}[t]
\vspace{-2em}
  \centering
\includegraphics[width=0.98\linewidth]{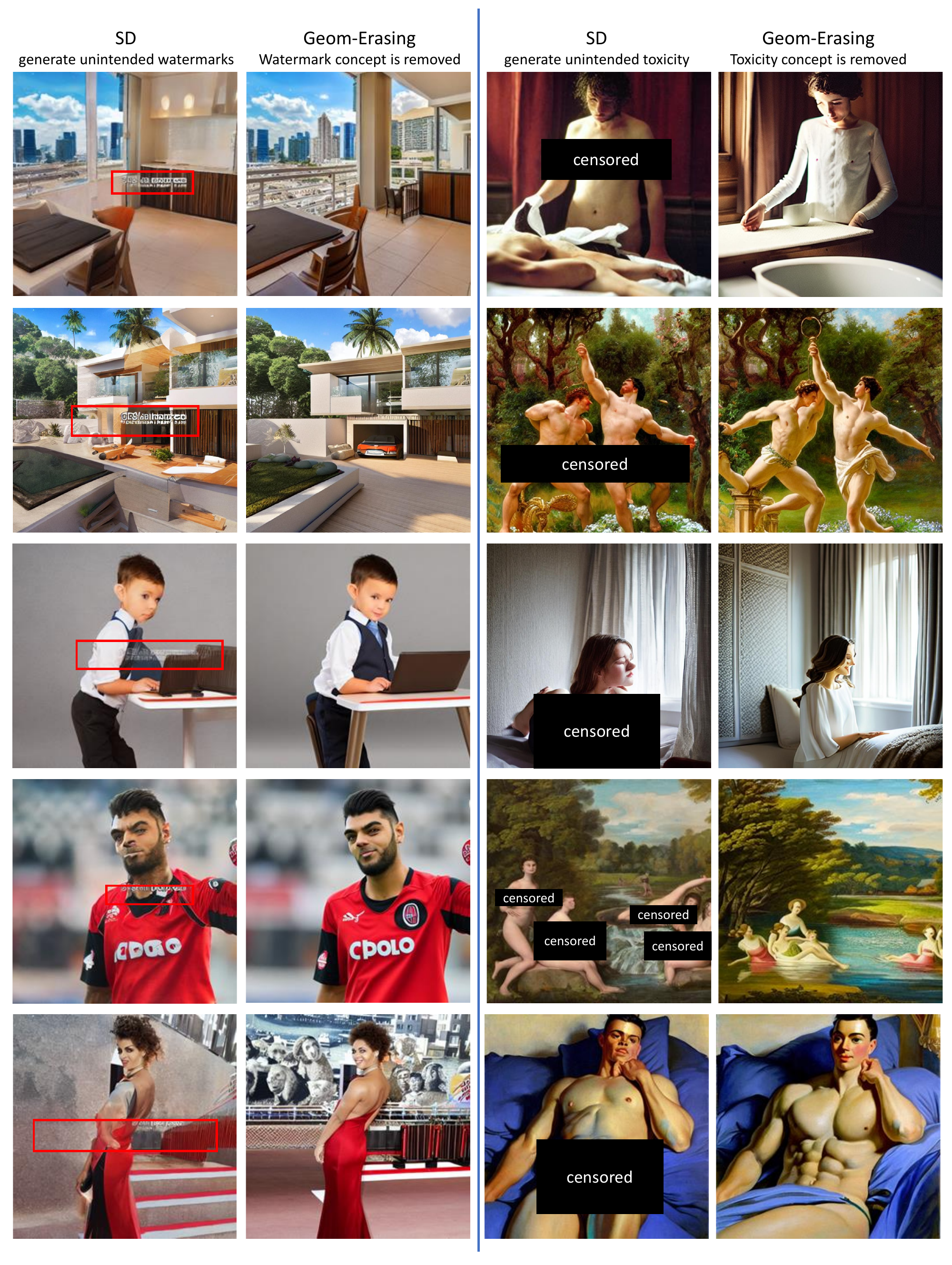}
\vspace{-1em}
  \caption{\textbf{Erasing Implicit Concepts from SD.} We successfully remove watermark and toxicity concepts from generated images while retaining other contents. Optimal viewing is recommended in color and at an enlarged scale.}
  \label{fig:erase_sd}
\end{figure*}

\begin{figure*}[t]
  \centering
  \includegraphics[width=\linewidth]{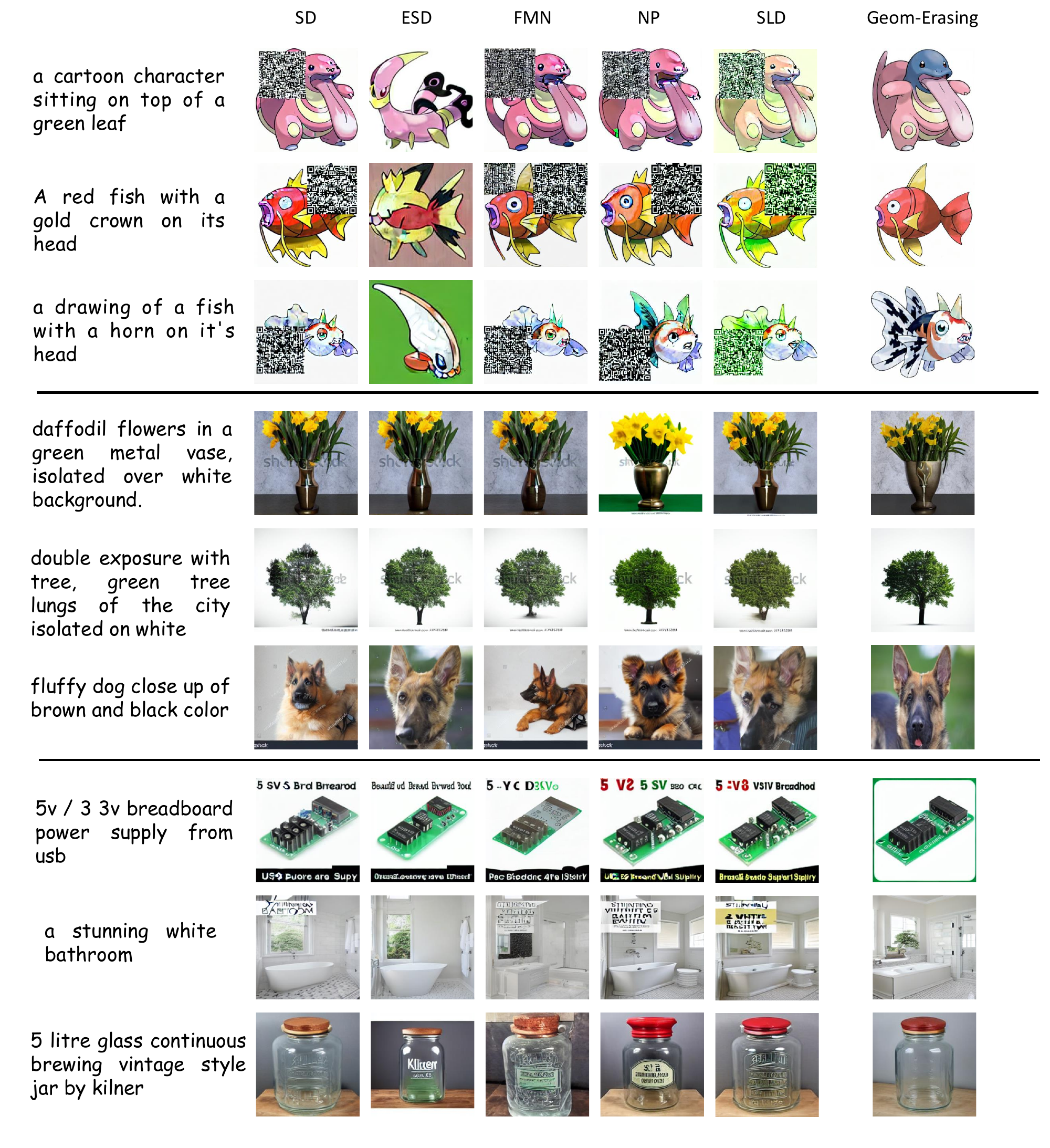}
  \caption{\textbf{Erasing implicit concept in ICD.}
  The first group of images are fine-tuned on ICD-QR. The middle and the bottom are fine-tuned on ICD-watermark and ICD-Text, respectively.}
  \label{fig:sota_more}
\end{figure*}

We present visualizations of implicit concept erasure under two settings; refer to Fig.~\ref{fig:erase_sd} for the first setting and Fig.~\ref{fig:sota_more} for the second setting. Notably, \methodname effectively removes ``watermark'' concepts in the original stable diffusion, including those challenging for human recognition. Intriguingly, for the ``toxicity'' concept, \methodname autonomously adds clothing to nude bodies, eliminating sexual content in the original images, while preserving other image contents. 
In the ICD erasure, specific implicit concepts are successfully eliminated, even within datasets containing undesired concepts. Both settings attest to the efficacy of \methodname.

\section{More Discussion}
\paragraph{Limitation and future work.}
\methodname proposes a novel geometric-aware framework~\cite{chen2023integrating}, originally designed for object detection data generation~\cite{han2021soda10m,li2022coda}, for the implicit concept removal in diffusion models.
Although effective, \methodname still relies on external specialized concept detectors to provide reasonable geometric information, 
suggesting that utilization of more general localizers (\eg, the activation maps of generative pre-training~\cite{chen2023mixed,zhili2023task} and contrastive 
learning~\cite{chen2021multisiam,liu2022task,zhang2021unleashing,zhang2022self}) would be an appealing future research direction.
Moreover, it is also interesting to utilize the geometric controls to remove harmful concepts in large language models~\cite{touvron2023llama,chen2023gaining,gou2023mixture}.

\end{document}